\documentclass{article}
\usepackage{spconf,amsmath,graphicx}

\usepackage{graphicx}

\usepackage{epsf}

\usepackage{amssymb}
\usepackage{amsmath}
\usepackage{latexsym}
\usepackage{comment}
\usepackage{graphicx}
\usepackage{comment}
\usepackage{textcomp}
\usepackage{float}
\usepackage[locale=FR]{siunitx}
\usepackage{subfigure}
\usepackage{setspace}
\usepackage{comment}
\usepackage{hyperref}
\usepackage{subfiles}

\usepackage[subtle]{savetrees}

\usepackage[dvipsnames]{xcolor}
\definecolor{gray}{RGB}{219, 48, 122}

\def\lf{\left\lfloor}   
\def\rf{\right\rfloor}

\usepackage{tikz}
\usetikzlibrary{calc}
\usetikzlibrary{positioning}
\tikzset{
    >=stealth,
    hair lines/.style={line width = 0.05pt, lightgray},
    true scale/.style={scale=#1, every node/.style={transform shape}},
}

\title{A Data Efficient End-To-End Spoken Language Understanding Architecture}
\name{Marco Dinarelli$^{\star}$ \qquad Nikita Kapoor$^{\star}$ \qquad Bassam Jabaian$^{\dagger}$ \qquad Laurent Besacier$^{\star}$}

			\address{$^{\star}$ LIG - Universit\'e Grenoble Alpes, France $^{\dagger}$LIA- Avignon Universit\'e, France}

\begin{document}

\maketitle

\begin{abstract}
%In this paper we present our first study of an end-to-end SLU system.
%Our models learn jointly acoustic and linguistic-sequential features on a complex semantic chunking and tagging task.
%Compared to previous work on the same data, our system \emph{i)} is based on a sequence-to-sequence model without attention and does not use any external language model, \emph{ii)} while trained on far less data than models evaluated on the same data, it provides competitive results, improving in several cases state-of-the-art models. 

End-to-end architectures have been recently proposed for spoken language understanding (SLU) and semantic parsing. Based on a large amount of data, those models learn jointly acoustic and linguistic-sequential features. Such architectures give very good results in the context of domain, intent and slot detection, their application in a more complex semantic chunking and tagging task is less easy. For that, in many cases, models are combined with an external a language model to enhance their performance. 

In this paper we introduce
a data efficient system which is trained  end-to-end, with no additional, pre-trained external module. One key feature of our approach is an incremental training procedure where acoustic, language and semantic models are trained sequentially one after the other.
The proposed model has a reasonable size and achieves competitive results with respect to state-of-the-art while using a small training dataset. 
In particular, we reach 24.02\% Concept Error Rate (CER) on MEDIA/test  while training on MEDIA/train  without any additional data.

%In this paper, we present a data efficient end-to-end SLU system based on a sequence-to-sequence model for semantic chunking without any attention mechanisms and without using any external language model. Our experimental study shows that this model, trained on a small amount of date compared to the one used to train the state-of-the-art systems, provides competitive performance.

\end{abstract}
\section{Introduction}
\label{sec:intro}

Spoken Language Understanding (SLU) aims at extracting a semantic representation from a speech signal in human-computer interaction applications \cite{DeMori1997:SDBook}. First SLU systems were based on \textit{pipeline} architectures where an automatic speech recognition (ASR) module generates a transcription of utterances and a SLU module predicts the semantic labels.
\textit{Pipeline} systems now tend to be replaced by \textit{end-to-end}\footnote{Our approach, like previous approaches in the literature, is end-to-end at inference time, that is we do not use any intermediate representation between speech and semantic level at decoding; however we do use transcriptions at training time.} architectures based on neural models, where semantic representations are produced directly from a speech input without using transcriptions \cite{DBLP:journals/corr/abs-1802-08395,DBLP:journals/corr/abs-1906-07601,ghannay:hal-01987740,DBLP:journals/corr/abs-1904-03670}.
Most of recently proposed end-to-end  models are based on sequence-to-sequence architectures. They were initially applied to speech translation \cite{berard-nips2016,weiss2017sequence} and then to SLU tasks where the main goal is to extract the domain and user intent from an utterance, together with some semantic slots \cite{DBLP:journals/corr/abs-1802-08395,DBLP:journals/corr/abs-1904-03670}.

In this paper we address end-to-end
semantic chunking and tagging of spoken utterances.
The most relevant works of the literature with respect to this task \cite{DBLP:journals/corr/abs-1906-07601,ghannay:hal-01987740} propose models based on Feed-Forward Neural Networks (FFNN) similar to the Deep Speech 2 model proposed for ASR \cite{DBLP-journals/corr/AmodeiABCCCCCCD15}, and an independent pre-trained language model re-scores semantic outputs.
Except for \cite{DBLP:journals/corr/abs-1904-03670}, most end-to-end SLU systems of the literature are trained on huge amount of data. \cite{DBLP:journals/corr/abs-1906-07601} also apply  pre-training and transfer learning %from tasks similar to SLU, namely Named Entity Recognition (NER).
from other NLP tasks such as Named Entity Recognition (NER).
%\laurent{[LB: I don't understand this sentence... transfer learning is made from other (not similar) tasks ?]}
% BJ {replaced by : also apply  pre-training and transfer learning for other NLP tasks such as Named Entity Recognition (NER)}

The contribution of this paper lies in the proposal of a data efficient architecture which is trained  end-to-end, with no additional pre-trained external module.
The proposed model achieves  competitive results with respect to state-of-the-art  while using a small training dataset (French MEDIA \cite{Bonneau-Maynard2005:media}) and having a reasonable computational footprint. %We believe this  makes it promising for embedded and private by design voice platforms \cite{DBLP:journals/corr/abs-1805-10190}.
In particular, we reach 24.02\% Concept Error Rate (CER) on MEDIA/test  while training on MEDIA/train  without any additional data.

%We evaluate our models on the French MEDIA corpus \cite{Bonneau-Maynard2005:media}, making our results comparable to latest state-of-the-art end-to-end systems, which are trained on much more data.
%While our results do not always improve the state-of-the-art, our models are much better in comparable training conditions. Our results compare also positively, or even outperform results of end-to-end systems trained on much more data than what we use in this work.

The remainder of this paper is organised as follows. After presenting the task addressed by this work in Section~\ref{sec:SLU}, we describe our sequence-to-sequence neural model in Section~\ref{sec:seq2seq}. Section~\ref{sec:eval} provides our experimental study on the French MEDIA corpus  and we conclude in Section~\ref{sec:conclusions}.

%\section{Spoken Language Understanding}
\section{SLU Task Addressed (MEDIA)}

\label{sec:SLU}

In this work we are interested in the task of semantic chunking and tagging of speech signals, corresponding to the user utterances in a conversation with a spoken dialog system. We focus on the specific domain of hotel information and reservation via an automatic system. This particular context is offered by the French  MEDIA corpus \cite{Bonneau-Maynard2005:media}.
It is made of $1~250$ human-machine dialogs acquired with a \textit{Wizard-of-OZ} approach, where \num{250} users followed \num{5} different reservation scenarios.
Spoken data was manually transcribed and annotated with domain concepts, following a rich ontology.
Statistics on the training, development and test data of the MEDIA corpus are shown in Table~\ref{tab:MEDIAStats}.
We note that, while all turns have been manually transcribed (\emph{total duration} in the table) and can be used to train ASR models, only user turns have been annotated with concepts (\emph{user sentences}) and can be used to train SLU models.

\begin{table}[t]
\begin{minipage}{1.0\linewidth}
    \centering
    \scriptsize
    \begin{tabular}{|l|rr|rr|rr|}
      \hline
      & \multicolumn{2}{|c|}{Train} & \multicolumn{2}{|c|}{Dev} & \multicolumn{2}{|c|}{Test}\\
      \hline
      total duration     &\multicolumn{2}{|c|}{41.5 hours} &\multicolumn{2}{|c|}{3.5 hours}&\multicolumn{2}{|c|}{11.3 hours} \\
      \# user sentences     &\multicolumn{2}{|c|}{12,908} &\multicolumn{2}{|c|}{1,259}&\multicolumn{2}{|c|}{3,005} \\
      \hline
      \hline
      & Words & Tags &  Words & Tags & Words & Tags \\
      \hline
      \# tokens	& 94,466 & 43,078 & 10,849 & 4,705 & 25,606 & 11,383 \\
      \# types		&  2,210 &     99 &    838 &    66 &  1,276 &     78 \\
      OOV\%	& --     & --     &  1.33  & 0.02  &  1.39  &  0.04  \\
      \hline
    \end{tabular}
    \caption{Statistics of the MEDIA corpus}
  \label{tab:MEDIAStats}
  \end{minipage}
\end{table}

\begin{figure}
	\includegraphics[width=1.0\linewidth]{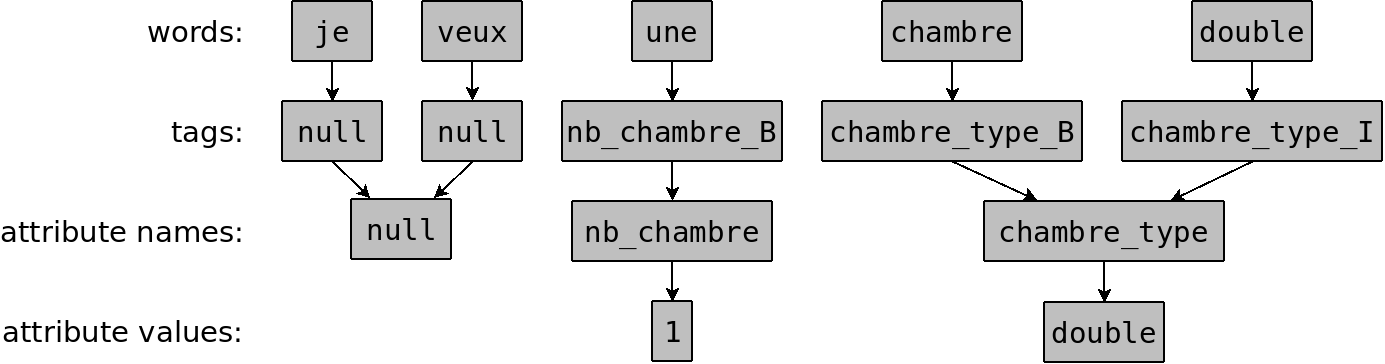}
	\caption{\small{Schema of concept (attribute names) and value (attribute values) extraction in the SLU task for spoken dialog systems \cite{Hahn.etAL-SLUJournal-2010}}}
	\label{fig:SLUSchema}
\end{figure}

Before the diffusion of neural networks, SLU was performed with pipeline systems \cite{dinarelli09:Interspeech,Hahn.etAL-SLUJournal-2010}. ASR was trained on large amount of data and refined on a specific SLU task. ASR transcripts were used as input to the SLU module, whose purpose was to tag words with the concepts. A further module was in charge to extract normalized values from tokens instantiating a particular concept. This is the processing applied also by a recent pipeline system based on neural networks \cite{DBLP:journals/corr/SimonnetGCEM17}.
The concept (attribute names) and value (attribute values) extraction schema is shown in Figure~\ref{fig:SLUSchema}.
In this work we focus on attribute names extraction only, and we decode directly whole concepts, without passing through the BIO intermediate format.

All these modules can be seen as different sub-tasks. Moreover the ASR can be further split into acoustic and language model learning sub-tasks. Thanks to neural networks, the sub-tasks can be learned jointly in an end-to-end framework.
The literature also shows that it might be convenient to learn incrementally the different sub-tasks involved in SLU: acoustic features, characters, tokens and finally concepts.
We adopt a similar \textit{incremental} strategy learning different features at different learning stages.
We use sequence-to-sequence neural models for learning jointly acoustic and linguistic features, without using any externally pre-trained language model to rescore local predictions from the acoustic feature encoder. Joint training is performed at different sub-task levels, eventually resulting in learning semantic features jointly with acoustic and linguistic features.
Our neural models are detailed in the following section.

\section{A data efficient Sequence-to-Sequence Model for SLU}
\label{sec:seq2seq}

\subsection{Basic, Sequential and 2-Stage Models}
\label{subsec:models}

The architecture proposed in this paper is based on sequence-to-sequence neural models \cite{Sutskever-2014-SSL-2969033.2969173,46201}.
The encoder has a similar architecture as the one used in the \emph{Deep Speech 2} architecture \cite{DBLP-journals/corr/AmodeiABCCCCCCD15}. It takes as input the spectrogram of the speech signal, which is passed through a stack of convolutional and recurrent layers, and generates representations of the output. This can be characters, tokens or semantic classes, depending on which sub-task is targeted.

The decoder has the same architecture as in \cite{DinarelliGrobol-Seq2BiseqTransformer-2019}. It has  characteristics of both recurrent and \emph{Transformer} \cite{46201} neural networks.
It takes as input the output of the encoder, but also its own previous predictions. These are integrated into the decoder as embeddings of discrete items (indexes), the hidden layer of the decoder embeds thus a concatenation of both acoustic and linguistic/semantic features.
Thanks to this choice, the architecture learns joint characteristics of both acoustic and language models, when characters or tokens are the items to be predicted. The model learns jointly acoustic and semantic features, when semantic tags are the items to be predicted. The fact that the current prediction depends also on previous predictions, allows the hidden layer to encode the sequential nature of output items.

%%%%%%%%%%%%%%   Basic Model   %%%%%%%%%%%%%%
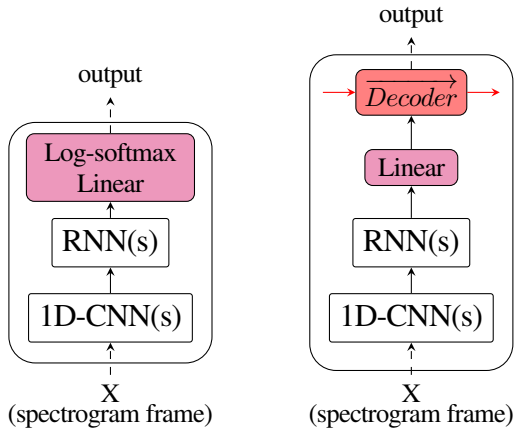
\begin{figure}
\center
\begin{tikzpicture}
	
	\begin{scope}[local bounding box=net]
	
	%% Components for the Basic Model
	\node (o) at (0,5.2) {output};
	
	\node[draw, rectangle, text width=2.0cm, text centered, rounded corners=3pt, fill=gray!50] (a2) at (0,4) {Log-softmax Linear};
	
	\node[draw, rectangle, rounded corners=1pt, scale=1.2] (t) at (0,3) {RNN(s)};
	
	%\node[draw, rectangle, rounded corners=3pt, fill=gray!50] (a1) at (0,2) {Layer Normalization};
	
	\node[draw, rectangle, rounded corners=1pt, scale=1.2] (c) at (0,2) {1D-CNN(s)};
	
    	\node (i) at (0,1.0) {X};
    	\node (i_L) at (0, 0.7) {(spectrogram frame)};
    
    %% Components for the Sequential Model
    \node (seqo) at (4,6.0) {output};
    
    \node[draw, rectangle, rounded corners=3pt, fill=red!50] (fwDec) at (4,5) {$\overrightarrow{Decoder}$};
    
    \node[draw, rectangle, text width=1.0cm, text centered, rounded corners=3pt, fill=gray!50] (seqa2) at (4,4) {Linear};
    
    \node[draw, rectangle, rounded corners=1pt, scale=1.2] (seqt) at (4,3) {RNN(s)};
    
    \node[draw, rectangle, rounded corners=1pt, scale=1.2] (seqc) at (4,2) {1D-CNN(s)};
    
    \node (seqi) at (4,1.0) {X};
    \node (seqi_L) at (4, 0.7) {(spectrogram frame)};
    
    %% Links for the Basic Model
    \draw[->, dashed] (i) -- (c);
    
    \draw[->] (c) -- (t);
    
    \draw[->] (t) -- (a2);
    
    \draw[->, dashed] (a2) -- (o);

	%% Frame for the Basic Model
    \tikzstyle{noeud}=[minimum width=2.7cm,minimum height=3.2cm, rectangle,rounded corners=10pt,draw,text=red,font=\bfseries]
    \node[noeud] (N) at (0,3.0) {};
    
    %% Links for the Seqeuntial Model
    \draw[->, dashed] (seqi) -- (seqc);
    
    \draw[->] (seqc) -- (seqt);
    
    \draw[->] (seqt) -- (seqa2);
    
    \draw[->] (seqa2) -- (fwDec);
    
    \draw[->,dashed] (fwDec) -- (seqo);
    
    %% Frame for the Sequential Model
    \tikzstyle{noeud}=[minimum width=2.7cm,minimum height=4.2cm, rectangle,rounded corners=10pt,draw,text=red,font=\bfseries]
    \node[noeud] (seqN) at (4,3.4) {};
    
    %% Entering and leaving links for the Sequential Model
    \node[] (w0) at (2.7,5.0) {};
    \node[] (w1) at (5.3,5.0) {};
    \draw[red,->] (w0) to (fwDec.west);
    \draw[red,->] (fwDec.east) to (w1);

    \end{scope}

\end{tikzpicture}
    \caption{Basic Model (left), Sequential Model (right)}\label{fig:End2EndSLU_BasicModel}
\end{figure}

\textbf{Basic (acoustic) model.} Since in general it is easier to learn acoustic and linguistic-sequential features incrementally, we use the encoder of our architecture as a \emph{basic} model. In order to be trained individually, we add on top of it a linear layer with a \emph{log-softmax} output function.
A schema of this basic model is given in  Figure~\ref{fig:End2EndSLU_BasicModel} (left).

\textbf{Sequential (acoustic+language) model.} In order to obtain our sequence-to-sequence model, we replace the log-softmax with a decoder.
A schema of this sequential model is given in the Figure~\ref{fig:End2EndSLU_BasicModel} (right).
We note that the basic model predicts an output item for each spectrogram frame independently, as a sequence of local decisions. The sequential model in contrast takes previous predicted items into account for the current prediction, and thus makes  contextual decisions.

\textbf{2-stage (acoustic+language) model.} We use the same idea as \cite{DBLP:journals/corr/abs-1904-03670} who proposed a model learning phones and tokens together. %: a first model learns only phones and it is then used as first stage of a deeper 2-stage model decoding tokens.
Our only difference is that instead of predicting phones, we decode characters. This has the advantage of being pronunciation dictionary free.
%in order to derive the phone transcription for each token.
%Both tokens and characters can be extracted automatically from the manual transcriptions of the speech signal. 
Our 2-stage model is obtained by stacking a sequential model (cf. Figure~\ref{fig:End2EndSLU_BasicModel}, on the right) on top of a basic model (cf. figure~\ref{fig:End2EndSLU_BasicModel}, on the left).

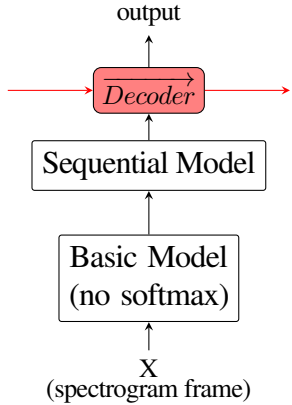
\begin{figure}
\center
\begin{tikzpicture}
	
	\begin{scope}[local bounding box=net]
    
    %% Components for the Sequential Model
    \node (seqo) at (4,6.0) {output};
    
    \node[draw, rectangle, rounded corners=3pt, fill=red!50] (fwDec) at (4,5) {$\overrightarrow{Decoder}$};
    
    \node[draw, rectangle, rounded corners=1pt, scale=1.2] (seqm) at (4,4) {Sequential Model};
    
    \node[draw, rectangle, text centered, text width=5.0em, rounded corners=1pt, scale=1.2] (basm) at (4,2.5) {Basic Model (no softmax)};
    
    \node (seqi) at (4,1.3) {X};
    \node (seqi_L) at (4, 1.0) {(spectrogram frame)};
    
    %% Links for the Seqeuntial Model
    \draw[->] (seqi) -- (basm);
    
    \draw[->] (basm) -- (seqm);
    
    \draw[->] (seqm) -- (fwDec);
    
    \draw[->] (fwDec) -- (seqo);
    
    %% Entering and leaving links for the Sequential Model
    \node[] (w0) at (2.0,5.0) {};
    \node[] (w1) at (6.0,5.0) {};
    \draw[red,->] (w0) to (fwDec.west);
    \draw[red,->] (fwDec.east) to (w1);

    \end{scope}

\end{tikzpicture}
    \caption{Schema of our End-to-End SLU model}\label{fig:SLUModel}
\end{figure}

\textbf{2-stage (acoustic+language+semantic) model.} The final SLU model is obtained by adding another decoder on top of a 2-stage model. It is trained to decode semantic concepts. A schema of our final SLU model is depicted in Figure~\ref{fig:SLUModel}.

\subsection{Incremental Training Strategy}
\label{subsec:training}

We use the neural architectures introduced in the previous section for training incrementally, one after the other, all the sub-tasks involved in SLU: acoustic features, characters, tokens and concepts decoding from speech.
%In other words, all models described in previous subsection are trained one after the other. 
We learn first a basic model for decoding characters, this is used as starting point for learning a sequential model for characters.
%The first learns acoustic features of characters only, while the second learns jointly a character language model together with acoustic features.
The sequential model for characters is used as starting point for a basic 2-stage model decoding tokens, which in turn initializes parameters of a sequential 2-stage model. The latter decodes tokens, it learns jointly acoustic and linguistic features of tokens, together with token sequences. It thus performs at the same time the role of acoustic and language models of traditional ASR systems.
Finally, a sequential model for decoding concepts (SLU) is learned by stacking a new decoder on top of a 2-stage model.
All models are learned minimizing the \emph{CTC} loss \cite{Graves:2006:CTC:1143844.1143891}.

When learning our sequential model with gold items, previous items given as input to the decoder are a much stronger predictor of the current item compared to representations of spectrogram input. The model gives thus much more importance to the previous items than to acoustic features, creating a mismatch between  training and testing conditions, when previous items must be predicted.
In order to avoid this behavior, we use a similar strategy as \cite{DBLP:journals/corr/BengioVJS15}: sequential models are trained starting with predicted items, when the learning rate is the greatest. After a given number of training epochs (an hyper-parameter), when weights have sufficiently been shaped from acoustic features, we switch to training with gold items. The rest of the training is lead by the error rate on development data (see section~\ref{subsec:settings}).
%If after a training epoch the error rate is not decreased, we switch again \laurent{[LB: what is switched ? still unclear...]} 

Another learning problem may be introduced by the very different length between input sequence (speech spectrograms) and gold output sequences (characters, tokens or concepts).
Let the input sequence have length $N$ and the output sequence have length $M$. In general $N \gg M$.\footnote{In our data we found that $\frac{N}{M} \leq 30$}
When the decoder is at processing step $i$, it has no information on which spectrogram frames to use as input.
This problem can be  solved using an attention mechanism \cite{DBLP-journals-corr-BahdanauCB14} to focus on the correct part of the input sequence depending on the part of the output sequence being decoded.
However, in this work we propose a simpler but efficient solution (inspired by \cite{Hochreiter-1997-LSTM,DBLP:journals/corr/JoulinM15,DBLP:journals/corr/abs-1805-04908}), based on a basic mechanism (since alignment is monotonic in SLU): we compute the ratio between output and input sequence lengths $r = \frac{M}{N}$, and when the model decodes at position $i$, it uses the sum of the encoder states around position $\lf i \cdot r \rf$.
%Noting that input-output sequence alignment is intrinsically monotonic in the SLU task, for this first study of an end-to-end SLU system we propose however a simple 
%This solution is based on previous works on algorithmic patterns and counting predictions of Recurrent Neural Networks like LSTMs \cite{Hochreiter-1997-LSTM,DBLP:journals/corr/JoulinM15,DBLP:journals/corr/abs-1805-04908}.

We further improve our training procedure with a variant of the curriculum strategy used in \cite{DBLP-journals/corr/AmodeiABCCCCCCD15}.
We sort speech turns based on their increasing length. Shorter turns, which have simpler sequential structures, are presented first to the model. After a given number of training epochs (an hyper-parameter), we switch to training with whole-dialog turn sequences.

\begin{table}[t!]
\centering
    \begin{tabular}{|l|c|c|}

        \hline
        \textbf{Model}  & \textbf{Dev WER} & \textbf{Test WER} \\
        \hline
        \hline
        
        \hline
        Basic Char			& 27.74 (*)	& -- \\
        Seq Char			& 24.05 (*)	& 23.32 (*) \\ \hline
        Basic Tok				& 30.79	& -- \\
        Seq Tok			& 29.42	& 28.71 \\ \hline
        Basic 2-Stage		& 28.00	& -- \\
        Seq 2-Stage	& \textbf{27.95}	& \textbf{27.01} \\ \hline
        \hline
        Seq 2-Stage (no-incremental)			& 63.70	& 63.61 \\
        Seq 2-Stage (no-curriculum)	& 28.15	& 27.96 \\
        \hline

    \end{tabular}
\caption{ASR Results on MEDIA - (*) is a character error rate}
\label{tab:ASRResults}
\end{table}

\section{Evaluation}
\label{sec:eval}

%The models presented in this work are trained on the MEDIA corpus described in section~\ref{sec:SLU}.

\subsection{Settings}
\label{subsec:settings}

The size of layers in our models, resulted from optimization on the Dev data, are as follows: the input spectrogram features are of dimension 81 and so is the dimension of convolutional layers, recurrent layers (LSTMs) have 256 dimensions. In the decoder, embeddings of previous predictions have 150 dimensions while hidden layers have 300. The decoder predicting concepts has twice more dimensions for each layer. Our basic and sequential models use only 1 CNN layer with stride 2 and 2 Bi-LSTM layers, in contrast to \cite{ghannay:hal-01987740,DBLP:journals/corr/abs-1906-07601} where 2 and 6 are used, respectively. \emph{Layer normalization} and \emph{Dropout} regularization \cite{JMLR:v15:srivastava14a} (with $p=0.5$) are applied between each two layers.
Our most complex model (see section~\ref{subsec:results}) has less than 9.8M parameters, in contrast to e.g. \cite{2018:EndtoEndSLU:Haghani.etal} with 97M.
All models are learned with an \emph{ADAM} optimizer \cite{AdamOpt:2015}, with learning rate of $0.0005$ decayed linearly over 60 epochs.
The training procedure starts with the incremental training strategy described in previous section and using predicted items. After 5 epochs we switch to gold items. At this point, each time the error rate is not improved on DEV data for 2 consecutive epochs, we switch between gold and predicted items learning. %\laurent{[LB: this part should be made consistent withv the one of section 3.2 and put only in ONE subsection...not spread in TWO subsections...]}

\begin{table}[t!]
\centering
    \begin{tabular}{|l|c|c|c|}

        \hline
        \textbf{Model}  & \textbf{Training} & \textbf{Dev CER} & \textbf{Test CER} \\
        & \textbf{Speech} &  & \\
        \hline
        \hline
        
        Seq 2-Stage + $\overrightarrow{Dec}$		& 41.5h & 28.11	& 27.52 \\
        Seq 2-Stage + $\overrightarrow{Dec}$ tune	& 41.5h & 28.18	& 27.35 \\
        Seq 2-Stage + $\overrightarrow{Dec}$ XT		& 41.5h & \textbf{23.39}	& \textbf{24.02} \\ \hline
        \hline
        \multicolumn{4}{|c|}{State-of-the-art Models} \\ \hline
        E2E SLU \cite{ghannay:hal-01987740} 					& ~300h & 30.1	& 27.0 \\
        E2E Baseline \cite{DBLP:journals/corr/abs-1906-07601}				& 41.5h & --		& 39.8 \\
        E2E SLU \cite{DBLP:journals/corr/abs-1906-07601}		& ~500h & --		& 23.7 \\
        E2E SLU + curr. \cite{DBLP:journals/corr/abs-1906-07601}  & ~500h & -- & 16.4 \\
        
        \hline

    \end{tabular}
\caption{SLU Results on MEDIA. For full comparison we report the best result of \cite{DBLP:journals/corr/abs-1906-07601} (16.4), which is obtained with a beam-search decoding, while the others are obtained, like our results, with greedy decoding.}
\label{tab:SLUResults}
\end{table}

\subsection{Results}
\label{subsec:results}

% From Yannick's SLT 2018 paper:
% 0. Corpora used:
% 0.1 ESTER 1 (73 hours of training data)
% 0.2 ESTER 2 ??? ("not annotated with named entities and not used in this study")
% 0.3 ETAPE (22 hours of training data)
% 0.4 Quaero (12 hours, in total ???)
%
% 1. "The training corpus is composed of the training sets of ESTER1, ETAPE and Quaero. DEV and TEST are composed of DEV an TEST sets of ESTER1&2 and ETAPE.
% All of that accounts for 160 hours of speech. So how it happens that for ASR they have 297.7 hours of speech ???
% 1. for ASR and NER pretraining: ESTER 73 hours for training, ETAPE 22 hours, QUAERO 12 hours, total 107 hours
% 2. ASR only is trained on roughly 500 hours of speech
%
% From Yannick's IS 2019 paper:
% Corpora used
% 1. for ASR: EPAC, ESTER2, ETAPE, QUAERO, REPERE ("All these corpora were manually transcribed...")
% ETAPE and QUAERO are manually annotated. All the other corpora were annotated automatically. Finally a NER system is trained on all data, manually or automatically annotated.
% 2. for SLU: MEDIA and PORT-MEDIA. PORT-MEDIA has 26 concepts in common with MEDIA.
% 

We evaluate both ASR (Word Error Rate) and SLU (Concept Error Rate) results on MEDIA corpus (Dev and Test). 
%Results are given in terms of Character/Word Error Rate (WER) when decoding characters or tokens (we use WER for both), and Concept Error Rate (CER) when decoding concepts.\footnote{CER is actually the exact same metric as WER, but it is computed aligning predicted and gold concepts, instead of words or characters.}
%In this work we focus on concept decoding only, we do not extract normalized values.\laurent{[LB: unclear what this means for a non SLU specialist...]}
% \marco{Solved: this explained right after the introduction of figure 1.}

\textbf{ASR} results are presented in Table~\ref{tab:ASRResults}. Together with the character and the 2-stage models, we show performance from a model decoding directly tokens.
We observe 
%can see, not surprisingly, 
that the sequential model outperforms the basic model, which does not use information of the output's sequential structure.
The 2-stage model always outperforms the token model which demonstrates that using pre-trained character models gives an advantage over training directly for decoding tokens.
Training incrementally the different stages of the model is the most effective choice: %by the fact that, 
training a 2-stage model from scratch (\emph{no-incremental} in the table), the error rate is much higher (over 60\%).
Finally, using the curriculum learning (sort speech turns based on their increasing length) proves also to be slightly beneficial: 1\% lower WER compared to a model trained without curriculum strategy (\emph{no-curriculum} in the table). 
%The 2-stage model uses a sequential character model as starting point, which is learned with the curriculum strategy. These are thus the most constraining conditions for showing that the curriculum strategy provides an advantage.
Our best results are competitive with previously published ASR performance on MEDIA:  ASR used in \cite{Hahn.etAL-SLUJournal-2010} had an error rate of 30.4 on Dev data, which we improve by a large margin. Our own ASR baseline based on an HMM-DNN model trained with \emph{Kaldi}\footnote{https://kaldi-asr.org/} reached an error rate of 25.1 on Dev. The best results shown in table~\ref{tab:ASRResults} are not too far, and they provide the advantage of being trained end-to-end without any external data nor language model. Better ASR performances were published lately on MEDIA 
%This holds also for the best results of the literature, which are 23.6 
\cite{DBLP:journals/corr/SimonnetGCEM17,DBLP:journals/corr/abs-1906-07601} but these were trained on up to 12 times more ASR training data. In particular \cite{DBLP:journals/corr/abs-1906-07601} trains the ASR part of the model with 4 different corpora, for a total of roughly 300 hours of speech. \cite{DBLP:journals/corr/abs-1906-07601} used 5 different corpora, accounting for 500 hours of speech. \cite{DBLP:journals/corr/SimonnetGCEM17} uses even more data. Our system is trained on MEDIA training data only, consisting of 41.5 hours of speech. A comparison of the amount of speech training data used for end-to-end ASR systems on MEDIA is given in the column \emph{Training Speech} of table~\ref{tab:SLUResults}.\footnote{The amount of speech data for training the SLU is not always detailed in those papers, it is generally smaller or equal to the amount for training ASR reported in the table.}
%\laurent{[LB: if possible can we be more precise and say exactly what was used for acoustic and LM training in both references ?]}

\textbf{SLU} performances are given in Table~\ref{tab:SLUResults}. Our results can be compared with some previous works \cite{ghannay:hal-01987740,DBLP:journals/corr/abs-1906-07601}. We note however that results reported in \cite{ghannay:hal-01987740,DBLP:journals/corr/abs-1906-07601} are obtained with models trained with much more data exploiting NER tasks with transfer learning. In particular \cite{DBLP:journals/corr/abs-1906-07601} uses 3 NER corpora for bootstrapping an end-to-end system. This is then fine-tuned on a first SLU corpus similar to MEDIA, and finally on MEDIA (see \cite{DBLP:journals/corr/abs-1906-07601} for details). For training our system, we note that only user turns are annotated with concepts, these account for 16.8 hours, that is less than half of the 41.5 hours of speech available, containing both machine and user turns.
%\laurent{[LB: is it only more ASR training data or more ASR+SLU training data or more ASR data + transfer learning ? ;;; this should be more precise here...]}.
The only result that is obtained in similar training conditions as ours, is the baseline model of \cite{DBLP:journals/corr/abs-1906-07601}. We can see that our model improves such baseline by a large margin, proving that learning jointly acoustic and linguistic-sequential features in an end-to-end framework is more effective than rescoring outputs with an independent language model.
More importantly, our results are comparable to the best results reported in \cite{ghannay:hal-01987740}. Once again these are obtained with models trained with much more data and a curriculum strategy via transfer learning among different corpora. This outcome highlight even more our data efficient training procedure.

Our SLU results in Table~\ref{tab:SLUResults} are obtained with a model decoding both concepts and tokens. This choice is imposed by the need of keeping track of which words instantiate a concept. Using the example in Figure~\ref{fig:SLUSchema}, the words \emph{``chambre double''} (double room) for instance, instantiate the concept \textbf{chambre-type}. Our model, similar to \cite{ghannay:hal-01987740}, generates the output \emph{$<$ chambre double \textbf{chambre-type} $>$}, which allows for attribute value extraction.\footnote{In this work we focus however on concept extraction only} This choice constrains the model to learn chunking and tagging at the same time, which is a much harder problem than just tagging \cite{DeMori1997:SDBook}.
%\laurent{LB: After this point, the explanations are difficult to follow/understand...They should be rephrased and more detailed i guess...]}
To improve this, we propose two alternatives: (1) refining token representations during  SLU model training (\emph{tune} in the table) and (2)
%In order to make training easier, we 
decoupling chunking and tagging using a second decoder which decodes only concepts (\emph{XT} in the table for \emph{extended}). In this latter decoding strategy, the first decoder generates a first output with concept boundary annotation while the  second decoder generates concepts only, aligned to the output of the previous decoder.
As we can see in the results, the \emph{XT} model obtains results comparable to those obtained by models trained with much more training data.
%proving that joint learning with sequence-to-sequence models is more suitable for end-to-end architectures.

\section{Conclusions}
\label{sec:conclusions}

In this paper we proposed a  data efficient end-to-end SLU system.
%sequence-to-sequence models for end-to-end spoken language understanding.
Our model learns jointly acoustic and linguistic-sequential features, allowing to train SLU models without an explicit language model module (pre-trained independently and/or on huge amount of data).
The efficiency of our system comes mostly from an incremental training procedure.
%While our results don't improve the absolute state-of-the-art, they compare positively to results obtained with models trained on much more data, and outperforming them in some cases.
The proposed model achieves  competitive results with respect to state-of-the-art  while using a small training dataset and having a reasonable computational footprint. 

\section{Acknowledgements}

This work was funded by the Institute Carnot Cognition.

%\vfill\pagebreak

% -------------------------------------------------------------------------
\bibliographystyle{IEEEbib}
\bibliography{icasspbiblio}

\end{document}